%% file: main.tex
\documentclass[letterpaper]{article} 
\usepackage{aaai23}  
\usepackage{times}  
\usepackage{helvet}  
\usepackage{courier}  
\usepackage[hyphens]{url}  
\usepackage{graphicx} 
\usepackage{natbib}  
\usepackage{caption} 
\usepackage{algorithm}
\usepackage{algorithmic}

\usepackage{newfloat}
\usepackage{listings}
\DeclareCaptionStyle{ruled}{labelfont=normalfont,labelsep=colon,strut=off} 
\floatstyle{ruled}
\newfloat{listing}{tb}{lst}{}
\floatname{listing}{Listing}
\title{Hierarchical Event Grounding}
\author{
    Jiefu Ou,
    Adithya Pratapa,
    Rishubh Gupta,
    Teruko Mitamura
}
\affiliations{
    Language Technologies Institute, Carnegie Mellon University\\
    \{jiefuo, vpratapa, rishubhg, teruko\}@andrew.cmu.edu
}

\usepackage{amsmath}
\usepackage{amssymb}
\usepackage{booktabs}
\usepackage{tikz}
\usepackage{subcaption}
\usepackage{graphicx}
\usepackage{pgfplots}
\pgfplotsset{compat=1.17}
\usetikzlibrary{positioning,shapes,decorations,automata,plotmarks,patterns}
\usepackage{todonotes}

\begin{document}

\maketitle

\begin{abstract}
Event grounding aims at linking mention references in text corpora to events from a knowledge base (KB). Previous work on this task focused primarily on linking to a single KB event, thereby overlooking the hierarchical aspects of events. Events in documents are typically described at various levels of spatio-temporal granularity \cite{glavas-etal-2014-hieve}. These hierarchical relations are utilized in downstream tasks of narrative understanding and schema construction. In this work, we present an extension to the event grounding task that requires tackling hierarchical event structures from the KB. Our proposed task involves linking a mention reference to a set of event labels from a subevent hierarchy in the KB. We propose a retrieval methodology that leverages event hierarchy through an auxiliary hierarchical loss \cite{murty-etal-2018-hierarchical}. On an automatically created multilingual dataset from Wikipedia and Wikidata, our experiments demonstrate the effectiveness of the hierarchical loss against retrieve and re-rank baselines \cite{wu-etal-2020-scalable,pratapa-etal-2022-multilingual}. Furthermore, we demonstrate the systems' ability to aid hierarchical discovery among unseen events.\footnote{Code is available at \url{https://github.com/JefferyO/Hierarchical-Event-Grounding}}
\end{abstract}

\section{Introduction}
Grounding entity and event references from documents to a large-scale knowledge base (KB) is an important component in the information extraction stack. While entity linking has been extensively explored in the literature \cite{ji-grishman-2011-knowledge}, event linking is relatively unexplored.\footnote{We interchangeably use the terms, grounding and linking.} Recently, \citet{pratapa-etal-2022-multilingual} presented a dataset for linking event references from Wikipedia and Wikinews articles to Wikidata KB. However, this work limited the grounding task to a subset of events from Wikidata, missing out on hierarchical event structures available in Wikidata.

\input{images/hierarchical_linking.tex}

Text documents often describe events at varying levels of spatio-temporal granularity. Figure \ref{fig:hierarchical_linking} illustrates this through three text snippets from English Wikipedia. The mentions `Falaise Gap', `Normandy campaign', and `western campaign' refer to three separate events from Wikidata.\footnote{Mention is a text span that refers to an underlying event.} These three events (Q602744, Q8641370, Q216184) themselves constitute a hierarchical event structure.

Prior work studied hierarchical relations between events as a part of datasets and systems proposed for narrative understanding \cite{glavas-etal-2014-hieve}, event sequencing \cite{mitamura2017events} and schema construction \cite{du-etal-2022-resin}. These works focused on hierarchical relations between mentions (e.g., subevent). In this work, we instead focus on \emph{hierarchical relations between grounded mentions} (i.e., events in a KB). This allows for studying hierarchy at a coarser level and leverages information across numerous mentions. To this end, we extend \citet{pratapa-etal-2022-multilingual} to include hierarchical event structures from Wikidata (Figure \ref{fig:hierarchical_linking}). In contrast to prior work, our formulation captures the hierarchical aspects by including non-leaf events such as `Operation Overlord' and `Western Front (World War II)'. Our formulation presents a challenging variant of the event linking task by requiring systems to differentiate between mentions of child and parent events.

For the proposed hierarchical linking task, we present a baseline that adopts the retrieve \& re-rank paradigm, which has been previously developed for entity and event linking tasks \cite{wu-etal-2020-scalable, pratapa-etal-2022-multilingual}. To enhance the system with hierarchy information, we present a methodology to incorporate such information via a hierarchy-aware loss \cite{murty-etal-2018-hierarchical} during the retrieval training. We experiment with the proposed systems on a multilingual dataset. The dataset is constructed by collecting mentions from Wikipedia and Wikinews articles that link to a set of events in Wikidata. Experiments on the collected dataset show the effectiveness of explicitly modeling event hierarchies.

Finally, we present a method for zero-shot hierarchy discovery in Wikidata (\S\ref{ssec:hierarchical_rel_ex}). We obtain a score for potential child-parent relations between two Wikidata events by computing the fraction of overlapping mentions from text documents. Results on an unseen subset of event hierarchies illustrate the effectiveness of our linking system in discovering hierarchical structures. Our key contributions are,

\begin{itemize}
    \item We propose the hierarchical event grounding task which requires linking mentions from documents to a set of hierarchically related events in a KB.
    \item We collect a large-scale multilingual dataset for this task that consists of mentions from Wikipedia and Wikinews articles linking to a set of events from Wikidata that are organized into hierarchies.
    \item We present a methodology that incorporates hierarchy-based loss for the grounding task. We show improvements over competitive retrieve-and-rerank baselines.
    \item We demonstrate an application of our linking systems for zero-shot hierarchical relation extraction.
\end{itemize}

\section{Related Work}
\label{sec:related_work}

\paragraph{Event Linking:} \citet{nothman-etal-2012-event} proposed linking event references in newswire articles to an archive of first reports of the events. Recent work on this task focused on linking mentions to knowledge bases like Wikipedia \cite{yu-etal-2021-event} and Wikidata \cite{pratapa-etal-2022-multilingual}. Our work is built upon the latter with a specific focus on hierarchical event structures in Wikidata.

\paragraph{Event Typing:} Given an event mention span and its context, typing aims at classifying the mention into one or more types from a pre-defined ontology. Commonly used ontologies include ACE 2005 \cite{walker-etal-2006-ace}, Rich-ERE \cite{song-etal-2015-light}, TAC-KBP \cite{mitamura2017events}. In contrast, event linking grounds mentions to one or more events from a KB (e.g., World War II in Wikidata vs Conflict type in ACE 2005).

\paragraph{Relation Extraction:} Extracting temporal, causal, and sub-event relations has been an integral part of event extraction pipelines. \citet{glavas-etal-2014-hieve} presented a dataset for studying hierarchical relations among events in news articles. \citet{ning-etal-2018-multi} studied event temporal relations, and \citet{han-etal-2021-ester} proposed extracting event relations as a question-answering task.

\paragraph{Hierarchy Modeling:} \citet{chen-etal-2020-hierarchical} presented a rank-based loss that utilizes entity ontology for hierarchical entity typing task. \citet{murty-etal-2018-hierarchical} explored a complex structured loss and \citet{onoe-etal-2021-modeling} utilized box embeddings \cite{vilnis-etal-2018-probabilistic} to model hierarchical relations between entity types.

\section{Hierarchical Event Grounding}
\label{sec:hierarchical_event_grounding}

The task of grounding involves linking event references in text documents to the corresponding entries in a knowledge base \cite{chandu-etal-2021-grounding}. \citet{pratapa-etal-2022-multilingual} studied linking event references from Wikipedia and Wikinews articles to Wikidata items. However, they restrict the dictionary of Wikidata items to leaf events. This ignores important parent events such as `Operation Overlord' and `Western Front
(World War II)' from Figure \ref{fig:hierarchical_linking}. In our preliminary analysis, we observed a significant number of mentions that refer to the parent events, motivating us to expand the dictionary to include all events.

Modeling hierarchy relations between events has been extensively studied \cite{glavas-etal-2014-hieve, mitamura2017events, du-etal-2022-resin}. These works typically focus on hierarchical relations among mentions in a document. In contrast, we focus on hierarchical relations among events in a KB. At a high level, this can be viewed as a combination of coreference resolution and hierarchy relation extraction.

\subsection{Task Definition}
\label{ssec:task_definition}

Consider an event knowledge base ($K$) that constitutes a set of events ($E$) and their relations ($R$). Each event ($e_i \in E$) has an id, title, and description. The relation set ($R$) includes both temporal and hierarchical (parent-child) links between events. Given an input mention span $m$ from a text document, the task is to predict an \emph{unordered} subset of events ($E_m \subset E$). This set $E_m$ constitutes events within a hierarchy tree, from the leaf to the root of the tree.\footnote{In Figure \ref{fig:hierarchical_linking}, the leaf (or atomic) events for the mentions `Normandy campaign' and `Falaise Gap' are \textit{Operation Overlord} and \textit{Falaise pocket} respectively.} In Figure \ref{fig:hierarchical_linking}, the mention `Normandy campaign' is to linked to the set \{Q8641370, Q216184\}, whereas the mention `Falaise Gap' is linked to \{Q602744, Q8641370, Q216184\}.

We follow prior work on linking \cite{logeswaran-etal-2019-zero} to formulate the task in a zero-shot fashion. Specifically, the set of event hierarchies for evaluation is completely unseen during training. Following \citet{pratapa-etal-2022-multilingual}, we present two task variants, 1. \emph{Multilingual}, where the event title and description are given in the same language as the mention and its context, and 2. \emph{Crosslingual}, where the event title and description are in English.\label{para:zero-mult}

An alternate task formulation involves traditional event linking followed by hierarchy propagation in the KB. However, such a formulation requires access to gold hierarchy relations at test time. In contrast, we present a task that facilitates hierarchy relation extraction among unseen events.

\subsection{Hierarchical Relation Extraction}
\label{ssec:hierarchical_rel_ex}

In addition to our key focus task of event linking, we explore hierarchical relation extraction for events. Similar to standard KB relation extraction \cite{trouillon-etal-2016-complex}, this involves predicting parent-child relationships in the KB. Specifically, given a hierarchical triple $(e_c, r, e_p)$ in $K$, where $e_c$ is the child of $e_p$ and $r$ is the child $\rightarrow$ parent relation, we mask $e_p$ and task models to retrieve it from the pool of events in $K$. To this end, we present a methodology to utilize our trained event-linking system for hierarchical relation extraction in Wikidata (\S\ref{ssec:hierarchical_rel_ex_methodology}).

\section{Dataset}
\label{sec:dataset}

To the best of our knowledge, there are no existing datasets for the task of hierarchical event linking. Therefore, we expand the XLEL-WD dataset \cite{pratapa-etal-2022-multilingual} to include hierarchical event structures.

\subsection{Event and Mention Collection}
\label{ssec:event_mention_collection}

Following prior work, we use Wikidata as our KB and follow a three-step process to collect events and their mentions. First, events are identified from Wikidata items by determining whether they have caused a change of state and possess spatio-temporal attributes. Then each event is associated with a set of language Wikipedia articles following the pointer contained in the event Wikidata item page. The title and description for each event in different languages are therefore obtained by taking the title and first paragraph of the associated language Wikipedia article. Finally, mentions linked to each event are collected by iterating over the language Wikipedia and identifying hyperlinks to the event-associated Wikipedia articles (obtained in the previous step). The anchor text of hyperlinks is taken as mentions and the surrounding paragraph of each mention is extracted as context.

\subsection{Hierarchy Construction}
\label{ssec:hierarchy_construction}

We further organize the collected pool of events into hierarchical trees by exploring property nodes from Wikidata. Property nodes act as edges between Q-nodes in Wikidata. We utilize two asymmetric and transitive properties, \emph{has-part} (P527) and \emph{part-of} (P361). Given two events, $e_1$ and $e_2$, if there exist edges such that $e_1$ has-part $e_2$ or $e2$ part-of $e_1$, we mark $e_1$ as the parent event $e_2$ and add the edge ($e2$, part-of, $e_1$) into the hierarchies. The full hierarchies are yielded by following such procedure and iterating over the full set of candidate events collected in \S\ref{ssec:event_mention_collection}. 

\input{images/wikidata_hierarchy.tex}

\subsection{Zero-shot Setup}
\label{ssec:zero_shot_splits}

For zero-shot evaluation, the train and evaluation splits should use disjoint hierarchical event trees from Wikidata. However, just isolating trees might not be sufficient. As shown in Figure \ref{fig:wikidata_hierarchy}, event trees can be a part of a larger temporal sequence. For instance, the events `2016 Summer Olympics', and `2020 Summer Olympics' share very similar hierarchical structure. To overcome this issue, we instead split events based on connected components of both hierarchical and temporal relations. In particular, candidate events are first organized into connected components that are grown by the two hierarchical properties: \textit{has-part} and \textit{part of} and two temporal properties: \textit{follows} (P155) and \textit{followed-by} (P156). The connected components are then assigned to disjoint splits.

After building the hierarchies among events, the final dictionary includes only events that are part of a hierarchy (event tree) of height $\geq$ 1. However, to conduct realistic evaluations that do not assume any prior knowledge on whether events belong to any hierarchy, all the events collected in \S\ref{ssec:event_mention_collection}, no matter whether they are part of a hierarchy or not, are presented to models as candidates at inference time.

\subsection{Wikinews Evaluation}
\label{ssec:wikinews_eval_set}

In addition to Wikipedia, assessing whether a system can generalize to new domains (e.g., news reports) has been regarded as a vital evaluation for entity \cite{botha-etal-2020-entity} and event linking systems \cite{pratapa-etal-2022-multilingual}. We follow the same procedure described above to construct a test set based on Wikinews articles with hyperlinks to Wikidata. 

\subsection{Dataset Statistics}
\label{ssec:dataset_stats}

To this end, we automatically compile a dataset with event hierarchies of maximum height$=$3 that consists of 2K events and 937K across 42 languages. The detailed statistics of the train-dev-test split as well as the Wikinews evaluation set (WN) are presented in Table \ref{tab:dataset}. Other than the events within hierarchies, we use the full set of events (including singletons) as the pool of candidates, this expands the effective size of our event dictionary to nearly 13k.

We perform a human evaluation of the parent-child relations in the development split of our dataset.\footnote{Two authors on this paper independently annotated the relations, followed by an adjudication phase to resolve disagreements.} We find the accuracy to be 96.7\%, highlighting the quality of our hierarchical event structures.

\input{tables/dataset}

\section{Methodology}

\label{sec:methodology}

\subsection{Baseline}
\label{ssec:baseline}

We use the standard retrieve and re-rank approach \cite{wu-etal-2020-scalable} as our baseline,

\paragraph{Retrieve:} We use two multilingual bi-encoders to independently encode mention (+context) and event candidates. We use the dot product between the two embeddings as the mention-event pair similarity score. To adapt this to our set-of-labels prediction task, for each mention $m$ and its target events set $E_m$, we pair $m$ with every event in $E_m$ to obtain $|E_m|$ mention-event pairs as positive samples. During each training step, the bi-encoder is optimized via the binary cross entropy loss (BCE) with in-batch negatives. At inference, for each mention, the bi-encoder retrieves the top-$k$ (e.g., $k=8$) events via nearest neighbor search.
 
\paragraph{Rerank:} We use a single multilingual cross-encoder to encode a concatenation of mention (+context) and event candidate. We follow prior work to pass the last-layer \texttt{[CLS]} through a prediction head to obtain scores for retrieved mention-event pairs. Due to computational constraints, the cross-encoder is trained only on the top-$k$ candidates retrieved by the bi-encoder. Cross-encoder is also optimized using a BCE loss that maximizes the score of gold events against other retrieved negatives for every mention. At inference, we output all the retrieved candidates with a score higher than a threshold ($\tau_c$; a hyperparameter) as the set of predicted events.

For both the bi-encoder and cross-encoder, we use XLM-RoBERTa \cite{conneau-etal-2020-unsupervised} as the multilingual transformer encoder.

\subsection{Encoding Hierarchy} 
\label{subsec:encode_hierarchy_info}

The baselines described above enable linking mentions to multiple KB events. However, they predict a flat set of events, overlooking any hierarchical relationships amongst them. To explicitly incorporate this hierarchy, we add a hierarchy aware loss in the bi-encoder training \cite{murty-etal-2018-hierarchical}. In addition to scoring mention-event pairs, the bi-encoder is also optimized to learn a scoring function $s(e_p, e_c)$ with the parent-child event pair $e_p, e_c \in \mathcal{K}$.

We parameterize $s$ based on the ComplEx embedding \cite{trouillon-etal-2016-complex}. It has been shown to be effective in scoring asymmetric, transitive relations such as hypernym in WordNet and hierarchical entity typing \cite{murty-etal-2018-hierarchical, chen-etal-2020-hierarchical}. ComplEx transforms type and relation embeddings into a complex space and scores the tuple with the real part of the Hermitian inner product. In our implementation, we use only the asymmetric portion of the product. 

In particular, given the embeddings of parent-child event pair $e_p, e_c \in \mathcal{K}$ as $\mathbf{e_p}, \mathbf{e_c} \in \mathbb{R}^d$ respectively, the score $s(e_p, e_c)$ is obtained by:

\begin{align}
    s(e_p, e_c) =& \langle \text{Im}(\mathbf{r}), \text{Re}(\mathbf{e_c}), \text{Im}(\mathbf{e_p}) \rangle \notag \\
                -& \langle \text{Im}(\mathbf{r}), \text{IM}(\mathbf{e_c}), \text{Re}(\mathbf{e_p}) \rangle \\
                =& \text{Im}(\mathbf{e_p}) \cdot (\text{Re}(\mathbf{e_c}) \odot \text{Im}(\mathbf{r})) \notag \\
                -& \text{Re}(\mathbf{e_p}) \cdot (\text{IM}(\mathbf{e_c}) \odot \text{Im}(\mathbf{r}))
\end{align}

Where $\odot$ is the element-wise product, $\mathbf{r} \in \mathbb{R}^d $ is a learnable relation embedding.

\begin{align}
    \text{Re}(\mathbf{e}) = W_{\text{Re}} \cdot \mathbf{e} + b_{\text{Re}}, \; \text{Im}(\mathbf{e}) = W_{\text{Im}} \cdot \mathbf{e} + b_{\text{Im}}
\end{align}

$\text{Re}(\mathbf{e})$ and $\text{Im}(\mathbf{e})$ are the biased linear projections of event embedding into real and imaginary parts of complex space respectively. $W_{\text{Re}}, W_{\text{Im}} \in \mathbb{R}^{d \times d}$ and $b_{\text{Re}}, b_{\text{Im}} \in \mathbb{R}^d$ are learnable weights. During training, a batch of $N_h$ parent-child event pairs is independently sampled and the bi-encoder is trained to minimize the in-batch BCE loss:
\begin{align}
    L_h = \frac{1}{N_h} \sum_{i=1}^{N_h} (-&\sum_{e_{cj} \in \mathcal{C}_i} \log(\sigma (s(e_{pi}, e_{cj}))) \notag \\ +& \sum_{e_{ck} \notin \mathcal{C}_i} \log(\sigma (s(e_{pi}, e_{ck}))))  
\end{align}
Where $\mathcal{C}_i$ denotes the set of children events in the batch that are the children of $e_{pi}$. We further explore three strategies for incorporating the hierarchy prediction in learning the bi-encoder:
\begin{itemize}
    \item \textbf{Pretraining}: the bi-encoder is pre-trained with the hierarchy-aware loss, followed by training with the mention linking loss.
    \item \textbf{Joint Learning}: in each epoch, the bi-encoder is jointly optimized with both the hierarchy-aware and mention linking loss.
    \item \textbf{Pretraining} + \textbf{Joint Learning}: bi-encoder is pretrained with the hierarchy-aware loss, followed by joint training with the hierarchy-aware and mention linking loss.
\end{itemize}

For each of the above bi-encoder configurations, we train a cross-encoder using the same training recipe as the baseline. We leave the development of hierarchy-aware cross-encoder models to future work.

\subsection{Hierarchical Relation Extraction}
\label{ssec:hierarchical_rel_ex_methodology}

In addition to the mention-linking, we propose a methodology to leverage the trained bi-encoders for hierarchical relation extraction. For each mention, we first retrieve the top-$k$ event candidates. We then construct a list of mentions ($M_e$) for each event $e$ in the dictionary. Finally, given a pair of events ($e_i, e_j$), we compute a score for potential child-parent ($e_i$--$e_j$) relation as follows,
\begin{align}
    h(e_i, e_j) = \frac{|M_{e_i} \cap M_{e_j}|}{|M_{e_i}|}
\end{align}
The scoring function $h$ is derived based on the intuition that if $e_j$ is the parent event of $e_i$, then all the mentions which are linked to $e_i$ should also be linked to $e_j$ by our linker. In such case, $M_{e_i}$ would be the subset of $M_{e_j}$ which indicates that $h(e_i, e_j) = 1$ approaches its maximum.

For each event, we iteratively calculate the child-parent score with every other event and rank them in descending order of the $h$ score. With this process, we could obtain a ranking of all other events as its candidate parents.
\section{Experiments}
\label{sec:experiments}

We experiment with the proposed configurations of bi-encoders and corresponding cross-encoders on the Wikipedia dataset for both multilingual and crosslingual tasks. To further assess out-of-domain generalization performance, we conduct the same experiments for baseline and the best-performing hierarchy-aware system on the Wikinews evaluation set. For the hierarchy relation extraction task, we evaluate the approach proposed in \S\ref{ssec:hierarchical_rel_ex_methodology} based on the retrieval results of the best-performing bi-encoder.

\subsection{Metrics}
\label{ssec:metrics}

\paragraph{Bi-encoder:} For bi-encoder, we follow prior work to report Recall@$k$ and extend it to multi-label version: measuring the fraction of mentions where all the gold events contained in the top-$k$ retrieved candidates. Since the longest path of hierarchies in the collected dataset consists of 4 events, we only evaluate with $k \geq 4$. And for $k < 4$, we instead report Recall@\emph{min}: the fraction of mentions where all the gold events contained in the top-$x$ retrieved candidates, where $x$ is the number of gold events for that mention. Recall@\emph{min} measures whether the bi-encoder could predict all and only the gold events with the minimal number of retrievals. For cases with single gold event, Recall@\emph{min} = Recall@1 which falls back to single event linking.

Our task requires the model to predict all the relevant events, from the atomic event to the root of the hierarchy tree. As an upper bound for model performance, we also report scores for predicting the most atomic event. In particular, the set of gold events is considered fully contained in top-$k$ retrieval of a mention if the most atomic gold event in the hierarchy is contained in the top-$k$ candidates. Our original task reduces to this atomic-only prediction task if the event hierarchy is known at test time. However, such an assumption might not be true in real-world settings. 

While Recall@$k$ is a strict and binary metric, i.e. the retrieval is counted as successful if and only if all the gold events are predicted, we further introduce a fraction version of it, denoted as Recall@$k$ (fraction), that allows for partial retrieval, with details in Appendix \ref{appendix:bi-encoder}.

\paragraph{Cross-encoder:} Similar to bi-encoder, we also follow previous work on entity linking to evaluate \emph{strict accuracy}, \emph{macro F1}, and \emph{micro F1} on the performance of cross-encoder. For a mention $m_i$, denote its gold events set as $E_i$, predicted events set as $\hat{E_i}$, with $N$ mentions:
\begin{align}
    \text{Strict Accuracy} &= \frac{\sum_{i=1}^N \mathbf{1}_{E_i = \hat{E_i}}}{N}
\end{align}
\begin{align}
    \text{MaP} = \frac{1}{N} \sum_{i=1}^N \frac{|E_i \cap \hat{E_i}|}{|\hat{E_i}|}, \; & \text{MaR} = \frac{1}{N} \sum_{i=1}^N \frac{|E_i \cap \hat{E_i}|}{|E_i|} \\
    \text{Macro F1} &= \frac{2\text{MaP}\cdot\text{MaR}}{\text{MaP} + \text{MaR}}
\end{align}
\begin{align}
    \text{MiP} = \frac{\sum_{i=1}^N |E_i \cap \hat{E_i}|}{\sum_{i=1}^N |\hat{E_i}|}, \; & \text{MiR} = \frac{\sum_{i=1}^N |E_i \cap \hat{E_i}|}{\sum_{i=1}^N |E_i|} \\
    \text{Micro F1} &= \frac{2\text{MiP}\cdot\text{MiR}}{\text{MiP} + \text{MiR}}
\end{align}

We additionally evaluate the strict accuracy of cross-encoders where we report the top-$x$ reranked candidates as predictions with $x$ being the number of gold events linked with the mention. This is the same condition as evaluating the bi-encoder on Recall@\emph{min}. It enables a direct comparison of strict accuracy to Recall@\emph{min} such that to assess if cross-encoders make improvements on bi-encoders. We denote the strict accuracy calculated under this condition as \emph{strict accuracy (top min)}. 

\paragraph{Hierarchical Relation Extraction:} As defined in \S\ref{ssec:hierarchical_rel_ex}, we evaluate the proposed hierarchical relation extraction method on whether it could identify the parent for a given child event. In particular, given an event with a ranked list of candidate parents (generated by the proposed method), we measure Recall@$k$ for the gold parent in the list. Since Recall@$k$ is ill-defined for events without a parent, we only calculate it for the non-root events within hierarchies of dev and test set. For those events that have parents but are not linked to any mentions by the bi-encoder, they are added as miss at every $k$. Such evaluation with Recall@$k$ measure is similar to the HIT@$k$ evaluation in the KB link prediction literature \cite{Bordes-etal-2011-LearningSE}.

\subsection{Bi-encoder Models}
\label{subsec:exp_bi-encoder_models}
As discussed in \S\ref{sec:methodology}, we evaluate the baseline bi-encoder (\emph{Baseline}) and three hierarchy-aware configurations: Hierarchy Pretraining (\emph{Baseline + HP}); Hierarchy Joint Learning (\emph{Baseline + HJL}); Hierarchy Pretraining \& Hierarchy Joint Learning (\emph{Baseline + HP + HJL}).

\subsection{Cross-encoder Models}
\label{subsec:cross_encoder_models_exp}

Given the top-$k$ retrieval results from each of the aforementioned bi-encoders, we train and evaluate a unique crossencoder respectively. The value of $k$ used for all cross-encoder experiments is selected to balance retrieval qualities (i.e.\ bi-encoder Recall@$k$ on dev set) and computation throughput (\S\ref{ssec:bi_res}). In case some of the gold events are not retrieved among top-$k$ candidates for the corresponding mention in the training set, we substitute each missing gold event for the negative candidates with the current lowest probability and repeat this process until all the missing gold events are added. At inference time, we apply a threshold $\tau_c$ to the reranked event candidates and emit those with score $\geq \tau_c$ as final predictions. If there is no event yielded, we add a \texttt{NULL} event to the prediction.

\textbf{Hierarchical Relation Extraction}: We apply the proposed method to top-4 retrieval results from the best-performing bi-encoder to perform hierarchical relation extraction. 

\section{Result and Analysis }
\label{sec:result_analysis}
\subsection{Bi-encoder}
\label{ssec:bi_res}
\input{plots_final/ml_bi_dev.tex}
\input{plots_final/xl_bi_dev.tex}
\input{tables/bi_cr_final}

Bi-encoder retrieval results on the dev split for both multi- and cross-lingual tasks are illustrated in Figure \ref{fig:ml_bi_dev} and Figure \ref{fig:xl_bi_dev} respectively. Since the gain in Recall@$k$ is relatively minor when doubling $k$ from 8 to 16 across all configurations and tasks, the cross-encoder is trained with the top 8 retrieved candidates, with the consideration of computation efficiency. It is also shown that all configurations attain better performance when evaluated by retrieving the most atomic event only (set of dense dots vs line plots), which reflects the benefits of following the gold hierarchies and indicates the performance upper-bound for current models that try to learn these hierarchies.

We further report the quantitative results of bi-encoder Recall@\emph{min} on dev and test set of both tasks in Table \ref{tab:bi_cr_final}. Among all the hierarchy-integration strategies, hierarchical pretraining offers consistent improvements on both tasks compared with the baseline. On the other hand, hierarchical joint learning presents a mixture of effects. In particular, it attains the best performance on the crosslingual test set when applied in conjunction with hierarchical pretraining while contributing negatively in all other scenarios.

In terms of task languages, all the multilingual configurations attain higher performance than their crosslingual counterparts, indicating that in general crosslingual task is more challenging than the multilingual task, which is similar to the single event linking scenario.

As described in Section \ref{ssec:metrics}, we further report bi-encoder results under Recall@$K$ (fraction) in Appendix \ref{appendix:bi-encoder}.

\subsection{Cross-encoder}
\label{ssec:cr_res}

Cross-encoder reranking results on both tasks are also shown in Table \ref{tab:bi_cr_final}. On the multilingual task, all the cross-encoders that are paired with hierarchy-aware bi-encoders outperform the baseline on strict accuracy and attain better or comparable performance on macro/micro F1. On the crosslingual task, (d) is the only hierarchy-aware system that outperforms the baseline across all metrics. All the models attain better results with Top Min accuracy and the relative performance differences between them remain similar to that of normal accuracy. Similar to the bi-encoder, the large performance gap of cross-encoders between the two tasks confirms that the crosslingual setting is more challenging.

\subsection{Bi-encoder vs.\ Cross-encoder}

We further investigate whether the cross-encoder could make improvements on its bi-encoder across all configurations. As discussed in \S\ref{ssec:metrics}, by comparing the strict accuracy of cross-encoders under the Top Min condition with the Recall@\emph{min} of associated bi-encoders, we find that cross-encoders further enhance bi-encoder performance on the test set in multilingual tasks while underperforms in other cases. For closer inspection into the performance of systems on each language, we report the per-language bi- and cross-encoder results in Table \ref{tab:ml_per_lang} and Table \ref{tab:xl_per_lang} in Appendix \ref{appendix:per-lang}.

\subsection{Hierarchy Discovery}
\label{ssec:hierarchy_discovery}

\input{tables/hd}
Table \ref{tab:hd} presents the hierarchical relation extraction results of our proposed set-based approach using the retrieved candidates by the best performance bi-encoder ((b) in Table \ref{tab:bi_cr_final}). On both tasks, the proposed method is able to assign high rankings to true parents for events within hierarchies, demonstrating its capability in aiding humans to discover new hierarchical relations on a set of previously-unseen events.
\subsection{Wikinews}
\input{tables/wn_final}
As shown in Table \ref{tab:bi_cr_wn_final}, applying our baseline and two of the hierarchy-aware linking systems ((c) in multilingual and (d) in crosslingual) on the Wikinews dataset results in a similar performance to that on Wikipedia mentions, which demonstrates that our methods could generalize well on the news domain.

\section{Conclusion \& Future Work}
\label{sec:conclusion}

In this paper, we present the task of hierarchical event grounding, for which we compile a multilingual dataset with Wikipedia and Wikidata. We propose a hierarchy-loss based methodology that improves upon a standard retrieve and re-rank baseline. Our experiments demonstrate the effectiveness of our approaches to model hierarchies among events in both multilingual and crosslingual settings. Additionally, we show promising results for zero-shot hierarchical relation extraction using the trained event linker. Some potential directions for future work include adapting encoders to directly include hierarchy and further exploring hierarchical relation extraction on standard datasets.

\appendix
\section*{Acknowledgments}
This material is based on research sponsored by the Air Force Research Laboratory under agreement number FA8750-19-2-0200. The U.S. Government is authorized to reproduce and distribute reprints for Governmental purposes notwithstanding any copyright notation thereon. The views and conclusions contained herein are those of the authors and should not be interpreted as necessarily representing the official policies or endorsements, either expressed or implied, of the Air Force Research Laboratory or the U.S. Government.

\bibliography{anthology,custom,aaai23}

\section{Computation Resources}

In our experiments, we use a single NVIDIA RTX A6000 GPU.

\section{Randomness}
We choose an arbitrary random seed and conduct all experiments with this seed.

\section{Hyper Parameters}
\subsection{Bi-Encoder}
\paragraph{Baseline} The hyper-parameters for baseline bi-encoder training are as follows
\begin{verbatim}
learning_rate 1e-05
num_train_epochs 10
schedule linear
warmup_proportion 0.05
max_context_length 128
max_cand_length 128
train_batch_size 64
eval_batch_size 64
bert_model xlm-roberta-base
type_optimization all_encoder_layers
shuffle
\end{verbatim}
Where \texttt{train\_batch\_size} is the batch size of mention-event pairs for optimizing the mention linking loss, \texttt{max\_context\_length} and \texttt{max\_cand\_length} control the length of context + mention and event title + description, respectively.

\paragraph{Hierarchical Pretraining (HP)} In the Hierarchical Pretraining, the additional hyper-parameters are as follows
\begin{verbatim}
struct_pretrain_epoch 0
struct_pretrain_batch_size 192
\end{verbatim}
Where \texttt{struct\_pretrain\_epoch 0} refers to pre-train bi-encoder with hierarchy-aware loss in the first epoch, and \texttt{struct\_batch\_size} is the batch size of parent-child event pairs for hierarchy-aware loss.

\paragraph{Hierarchical Joint Learning (HJL)} In the Hierarchical Joint Learning, the additional hyper-parameters are as follows
\begin{verbatim}
struct_loss_epoch 0
struct_loss_batch_size 128
struct_loss_wt 0.01
\end{verbatim}
Where \texttt{struct\_loss\_epoch 0} refers to train bi-encoder jointly with mention-linking and hierarchy-aware losses starting from the first epoch (and for all following epochs). The \texttt{struct\_loss\_wt} is the weight on hierarchy-aware loss when jointly optimized with mention linking loss (of which the weight $=1$)

\paragraph{HP + HJL} In the Hierarchical Pretraining \& Hierarchical Joint Learning, the additional hyper-parameters are as follow
\begin{verbatim}
struct_pretrain_epoch 0
struct_loss_epoch 1
struct_pretrain_batch_size 192
struct_loss_batch_size 128
struct_loss_wt 0.01
\end{verbatim}
Where the model is pretrained with hierarchical loss only for the first epoch and jointly optimized with the two losses in the following epochs.

\subsection{Cross-encoder}
\paragraph{Cross-encoder Training} In training cross-encoders, we specified the following hyper-parameters:
\begin{verbatim}
learning_rate 1e-05
num_train_epochs 5
schedule linear
warmup_proportion 0.05
max_context_length 128
max_cand_length 128
train_batch_size 16
eval_batch_size 16
bert_model xlm-roberta-base
type_optimization all_encoder_layers
add_linear
top_k 8
add_all_gold
warmup_proportion 0.1
\end{verbatim}
With \texttt{top\_k 8}, the cross-encoders are train with the top-8 retrieved candidates per mention from the bi-encoders.

\paragraph{Cross-encoder Inference}
At inference time, we select the threshold $\tau_c$ for emitting prediction from the set $\{0.001, 0.01, 0.1, 0.3, 0.5, 0.7, 0.9\}$, based on the performance on dev set. In terms of measuring the performance, we use the product of all three metrics, i.e. $\text{Strict Acc} \times \text{Macro F1} \times \text{Micro F1}$.

\section{Additional Results}
\subsection{Bi-encoder}
\label{appendix:bi-encoder}
\input{plots_final/ml_bi_dev_frac.tex}
\input{plots_final/xl_bi_dev_frac.tex}

In addition to Recall@$k$, which counts the retrieval at $k$ as successful if and only if all the gold events are predicted within the top-$k$ retrieval, we propose to also evaluate bi-encoders that measures partial retrieval. In particular, we introduce Recall@$k$ (fraction): for a given mention with $p$ gold events, of which $q$ are predicted in the top-$k$ retrievals, its Recall@$k$ (fraction) = $\frac{q}{p}$.

Figure \ref{fig:ml_bi_dev_frac} and figure \ref{fig:xl_bi_dev_frac} depict the bi-encoder retrieval performance under Recall@$k$ (fraction) on the dev split of multi- and crosslingual tasks, respectively. Compared with the corresponding performance on Recall@$k$ in figure \ref{fig:ml_bi_dev} and figure \ref{fig:xl_bi_dev}, every model attains a higher score under Recall@$k$ (fraction) while the relative performance gap remains similar.

\subsection{Per-language results}
\label{appendix:per-lang}
Table \ref{tab:ml_per_lang} and Table \ref{tab:xl_per_lang} display the bi-encoder and cross-encoder performance on each of the languages, on both multi- and crosslingual tasks. We report the best-performing bi-encoder + cross-encoder systems ((c) in multilingual and (d) in crosslingual task) by measuring $\text{Strict Acc} \times \text{Macro F1} \times \text{Micro F1}$. 
\input{tables/ml_per-lang}
\input{tables/xl_per-lang}

\end{document}

%% file: images/hierarchical_linking.tex
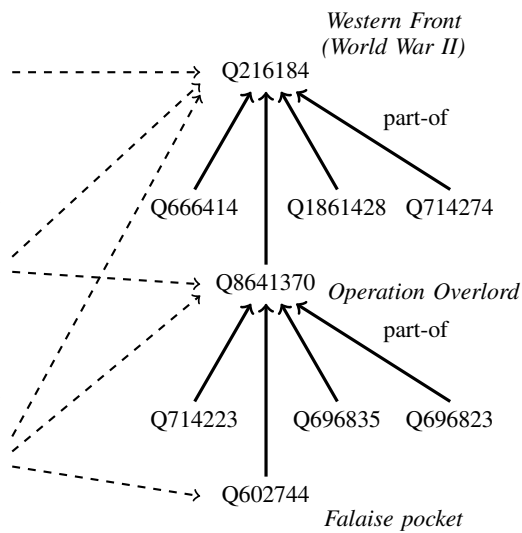
\begin{figure*}
\centering
\begin{tikzpicture}[
mentionnode/.style={font=\small,align=left,text width=10.5cm},
eventnode/.style={font=\small,align=center,text width=1.5cm},
descnode/.style={font=\small\itshape,align=center,text width=2.8cm},
edgenode/.style={font=\small,align=center}
]


\node[mentionnode] (mention_1) [
    ] {
        Gangl was promoted to Oberfeldwebel in November 1938. ... He returned to his regiment on May 14, 1940, and took part in the \textbf{western campaign}. There he served as the commander of a reconnaissance unit of the 25th Infantry Division of the Wehrmacht.
    };

\node[mentionnode] (mention_2) [
    below=0.3cm of mention_1
    ] {
        David Vivian Currie VC was awarded the Victoria Cross for his actions in command of a battle group of tanks from The South Alberta Regiment, artillery, and infantry of the Argyll and Sutherland Highlanders of Canada at St. Lambert-sur-Dives, during the final actions to close the Falaise Gap. This was the only Victoria Cross awarded to a Canadian soldier during the \textbf{Normandy campaign} (from 6 June 1944 to the end of August 1944)
    };

\node[mentionnode] (mention_3) [
    below=0.3cm of mention_2
    ] {
        David Vivian Currie VC was awarded the Victoria Cross for his actions in command of a battle group of tanks from The South Alberta Regiment, artillery, and infantry of the Argyll and Sutherland Highlanders of Canada at St. Lambert-sur-Dives, during the final actions to close the \textbf{Falaise Gap}. This was the only Victoria Cross awarded to a Canadian soldier during the Normandy campaign (from 6 June 1944 to the end of August 1944)
    };
    
\node[eventnode] (event_1) [
    right=2.5cm of mention_1
    ] {
        Q216184
    };
    
\node[eventnode] (event_2) [
    below=2.3cm of event_1
    ] {
        Q8641370
    };

\node[eventnode] (event_2_1) [
    below left=1.3cm and -0.8cm of event_1
    ] {
        Q666414
    };

\node[eventnode] (event_2_2) [
    below right=1.3cm and -0.8cm of event_1
    ] {
        Q1861428
    };

\node[eventnode] (event_2_3) [
    below right=1.3cm and 0.7cm of event_1
    ] {
        Q714274
    };
\node[eventnode] (event_3) [
    below=2.3cm of event_2
    ] {
        Q602744
    };

\node[eventnode] (event_3_1) [
    below left=1.3cm and -0.8cm of event_2
    ] {
        Q714223
    };

\node[eventnode] (event_3_2) [
    below right=1.3cm and -0.8cm of event_2
    ] {
        Q696835
    };

\node[eventnode] (event_3_3) [
    below right=1.3cm and 0.7cm of event_2
    ] {
        Q696823
    };

\node[descnode] (qnode_1) [
    above right=-0.2cm and -0.7cm of event_1
    ] {
        Western Front (World War II)
    };

\node[descnode] (qnode_2) [
    below right=-0.4cm and -0.3cm of event_2
    ] {
        Operation Overlord
    };

\node[descnode] (qnode_3) [
    below right=-0.2cm and -0.7cm of event_3
    ] {
        Falaise pocket
    };


\draw[->,very thick] (event_2.north) -- (event_1.south);
\draw[->,very thick] (event_2_1.north) -- ([xshift=-0.2cm]event_1.south);
\draw[->,very thick] (event_2_2.north) -- ([xshift=0.2cm]event_1.south);
\draw[->,very thick] (event_2_3.north) -- ([xshift=0.4cm]event_1.south) node[edgenode, midway, above right] {part-of};

\draw[->,very thick] (event_3.north) -- (event_2.south);
\draw[->,very thick] (event_3_1.north) -- ([xshift=-0.2cm]event_2.south);
\draw[->,very thick] (event_3_2.north) -- ([xshift=0.2cm]event_2.south);
\draw[->,very thick] (event_3_3.north) -- ([xshift=0.4cm]event_2.south) node[edgenode, midway, above right] {part-of};

\draw[->,black,thick,dashed] (mention_1.east) -- (event_1.west);
\draw[->,black,thick,dashed] ([yshift=-0.25cm]mention_2.east) -- ([yshift=-0.15cm]event_1.west);
\draw[->,black,thick,dashed] ([yshift=-0.45cm]mention_2.east) -- (event_2.west);
\draw[->,black,thick,dashed] ([yshift=-0.05cm]mention_3.east) -- ([yshift=-0.3cm]event_1.west);
\draw[->,black,thick,dashed] ([yshift=-0.25cm]mention_3.east) -- ([yshift=-0.15cm]event_2.west);
\draw[->,black,thick,dashed] ([yshift=-0.45cm]mention_3.east) -- (event_3.west);

\end{tikzpicture}
    \caption{An illustration of mention linking to hierarchical event structures in Wikidata. The left column shows three mentions (\textbf{highlighted}) with their contexts, and the right column presents a hierarchy of Q-nodes from Wikidata. Each mention is linked to a set of events from a hierarchy path from Wikidata (e.g., mention `Normandy campaign' is linked to a set of two events, \{Q8641370, Q216184\}).}
    \label{fig:hierarchical_linking}
\end{figure*}

%% file: images/wikidata_hierarchy.tex
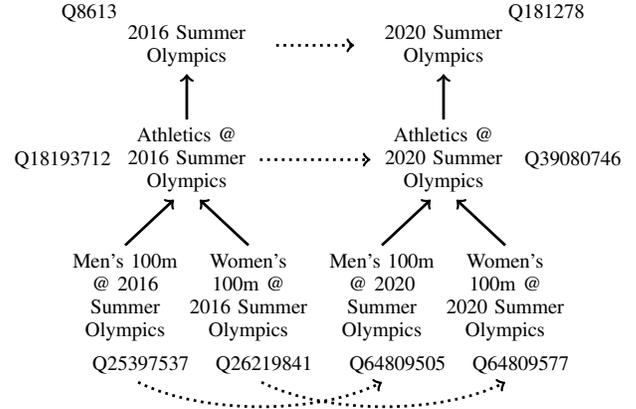
\begin{figure}
\centering
\resizebox{0.45\textwidth}{!}{
\begin{tikzpicture}[
eventnode/.style={font=\small,align=center,text width=2cm},
qnode/.style={font=\small,align=center,text width=1cm},
]

    
    
    

\node[eventnode] (node2016_1) [
    text width=2.5cm
    ] {
        2016 Summer Olympics
    };
\node[qnode] (qnode2016_1) [
    above left=-0.2cm and -0.5cm of node2016_1
    ] {
        Q8613
    };
    
\node[eventnode] (node2016_2) [
    below=0.7cm of node2016_1
    ] {
        Athletics @ 2016 Summer Olympics
    };
\node[qnode] (qnode2016_2) [
    left=0.4cm of node2016_2
    ] {
        Q18193712
    };
    
\node[eventnode] (node2016_3) [
    below left=0.7cm and -1.3cm of node2016_2
    ] {
        Men’s 100m @ 2016 Summer Olympics
    };
\node[qnode] (qnode2016_3) [
    below=0cm of node2016_3
    ] {
        Q25397537
    };
    
\node[eventnode] (node2016_4) [
    below right=0.7cm and -1.3cm of node2016_2
    ] {
        Women’s 100m @ 2016 Summer Olympics
    };
\node[qnode] (qnode2016_4) [
    below=0cm of node2016_4
    ] {
        Q26219841
    };

\node[eventnode] (node2020_1) [
    right=1.2cm of node2016_1,
    text width=2.5cm
    ] {
        2020 Summer Olympics
    };
\node[qnode] (qnode2020_1) [
    above right=-0.2cm and -0.5cm of node2020_1
    ] {
        Q181278
    };
    
\node[eventnode] (node2020_2) [
    below=0.7cm of node2020_1
    ] {
        Athletics @ 2020 Summer Olympics
    };
\node[qnode] (qnode2020_2) [
    right=0cm of node2020_2
    ] {
        Q39080746
    };

\node[eventnode] (node2020_3) [
    below left=0.7cm and -1.3cm of node2020_2
    ] {
        Men’s 100m @ 2020 Summer Olympics
    };
\node[qnode] (qnode2020_3) [
    below=0cm of node2020_3
    ] {
        Q64809505
    };
    
\node[eventnode] (node2020_4) [
    below right=0.7cm and -1.3cm of node2020_2
    ] {
        Women’s 100m @ 2020 Summer Olympics
    };
\node[qnode] (qnode2020_4) [
    below=0cm of node2020_4
    ] {
        Q64809577
    };


\draw[->,very thick] (node2016_2.north) -- (node2016_1.south);

\draw[->,very thick] (node2016_3.north) -- ([xshift=-0.2cm]node2016_2.south);

\draw[->,very thick] (node2016_4.north) -- ([xshift=0.2cm]node2016_2.south);






\draw[->,very thick] (node2020_2.north) -- (node2020_1.south);

\draw[->,very thick] (node2020_3.north) -- ([xshift=-0.2cm]node2020_2.south);

\draw[->,very thick] (node2020_4.north) -- ([xshift=0.2cm]node2020_2.south);




\draw[->,very thick,dotted] (node2016_1.east) -- (node2020_1.west);
\draw[->,very thick,dotted] (node2016_2.east) -- (node2020_2.west);
\draw[->,very thick,dotted] ([xshift=0.2cm]qnode2016_3.south) .. controls +(1,-0.5) and +(-1,-0.5) .. (qnode2020_3.south);
\draw[->,very thick,dotted] ([xshift=0.2cm]qnode2016_4.south) .. controls +(1,-0.5) and +(-1,-0.5) .. (qnode2020_4.south);

\end{tikzpicture}
}
    \caption{An illustration of hierarchical event structures in Wikidata. Each node represents an event from Wikidata. Solid arrows ({\protect\tikz[baseline] \protect\draw[->,very thick] (0pt, .5ex) -- (3ex, .5ex);}) and dotted arrows ({\protect\tikz[baseline] \protect\draw[->,very thick,dotted] (0pt, .5ex) -- (3ex, .5ex);}) denote hierarchical and temporal relations respectively.}
    \label{fig:wikidata_hierarchy}
\end{figure}

%% file: tables/dataset.tex
\begin{table}[t]
\centering
\begin{tabular}{@{}lrrrr@{}}
\toprule
                    & Train  & Dev   & Test & Wikinews \\
                    \midrule
\# mentions         & 751550 & 93047 & 91928 & 258  \\
\# events           & 2288   & 216   & 273   & 64   \\
\# trees            & 262    & 64    & 68    & 51     \\
\# children (avg.)  & 4.37   & 2.30  & 2.81  & 1.18   \\
tree depth (avg.)   & 1.23   & 1.03  & 1.06  & 0.22   \\
\bottomrule
\end{tabular}
\caption{Dataset statistics on train/dev/test splits from Wikipedia and Wikinews evaluation set. \# children (avg.) refers to the average number of children per non-terminal node. Due to its limited scale, there are only 200+ mentions in Wikinews articles that are linked to events within hierarchies. Therefore, for some of the event trees, there only exists mentions linking to the root node, which results in the average effective tree depth $< 1$.}
\label{tab:dataset}
\end{table}

%% file: plots_final/ml_bi_dev.tex
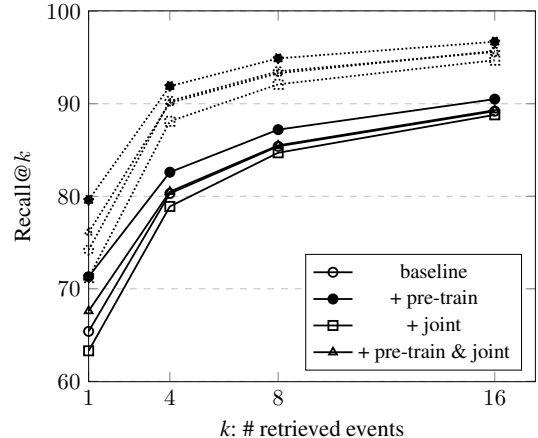
\begin{figure}[t]
\pgfplotstableread{plots_final/ml_bi_dev.tsv}\loadedtable
\resizebox{0.4\textwidth}{!}{
    \centering
    \begin{tikzpicture}
        \begin{axis}[
            xlabel={\textit{k}: \# retrieved events},
            ylabel={Recall@$k$},
            legend entries={baseline, + pre-train, + joint, + pre-train \& joint},
            legend pos=south east,
            legend style={font=\small},
            xmin=1,
            xtick={1,4,8,16},
            ymajorgrids=true,
            grid style=dashed,
            ]
        
        \addplot[mark=o,mark size=2,thick] table[x=k,y=b]
            {\loadedtable};
        \addplot[mark=*,mark size=2,thick] table[x=k,y=b-hp]
            {\loadedtable};
        \addplot[mark=square,mark size=2,thick] table[x=k,y=b-hjl]
            {\loadedtable};
        \addplot[mark=triangle,mark size=2,thick] table[x=k,y=b-hp-hjl]
            {\loadedtable};

        \addplot[mark=o,mark size=2,thick,densely dotted] table[x=k,y=atomic-b]
            {\loadedtable};
        \addplot[mark=*,mark size=2,thick,densely dotted] table[x=k,y=atomic-b-hp]
            {\loadedtable};
        \addplot[mark=square,mark size=2,thick,densely dotted] table[x=k,y=atomic-b-hjl]
            {\loadedtable};
        \addplot[mark=triangle,mark size=2,thick,densely dotted] table[x=k,y=atomic-b-hp-hjl]
            {\loadedtable};

        \end{axis}
    \end{tikzpicture}}
    \caption{Multilingual bi-encoder Recall@$k$ on the dev set. The densely dotted plots ({\protect\tikz[baseline] \protect\draw[-,thick,densely dotted] (0pt, .5ex) -- (3ex, .5ex);}) denote the prediction scores for the atomic label, an upper bound for model performance.}
    \label{fig:ml_bi_dev}
\end{figure}

%% file: plots_final/xl_bi_dev.tex
\begin{figure}[t]
\pgfplotstableread{plots_final/xl_bi_dev.tsv}\loadedtable
\resizebox{0.4\textwidth}{!}{
    \centering
    \begin{tikzpicture}
        \begin{axis}[
            xlabel={\textit{k}: \# retrieved events},
            ylabel={Recall@$k$},
            legend entries={baseline, + pre-train, + joint, + pre-train \& joint},
            legend pos=south east,
            legend style={font=\small},
            xmin=1,
            xtick={1,4,8,16},
            ymajorgrids=true,
            grid style=dashed,
            ]
        
        \addplot[mark=o,mark size=2,thick] table[x=k,y=b]
            {\loadedtable};
        \addplot[mark=*,mark size=2,thick] table[x=k,y=b-hp]
            {\loadedtable};
        \addplot[mark=square,mark size=2,thick] table[x=k,y=b-hjl]
            {\loadedtable};
        \addplot[mark=triangle,mark size=2,thick] table[x=k,y=b-hp-hjl]
            {\loadedtable};

        \addplot[mark=o,mark size=2,thick,densely dotted] table[x=k,y=atomic-b]
            {\loadedtable};
        \addplot[mark=*,mark size=2,thick,densely dotted] table[x=k,y=atomic-b-hp]
            {\loadedtable};
        \addplot[mark=square,mark size=2,thick,densely dotted] table[x=k,y=atomic-b-hjl]
            {\loadedtable};
        \addplot[mark=triangle,mark size=2,thick,densely dotted] table[x=k,y=atomic-b-hp-hjl]
            {\loadedtable};

        \end{axis}
    \end{tikzpicture}}
    \caption{Crosslingual bi-encoder Recall@$k$ on the dev set. The densely dotted plots ({\protect\tikz[baseline] \protect\draw[-,thick,densely dotted] (0pt, .5ex) -- (3ex, .5ex);}) denote the prediction scores for the atomic label, an upper bound for model performance.}
    \label{fig:xl_bi_dev}
\end{figure}
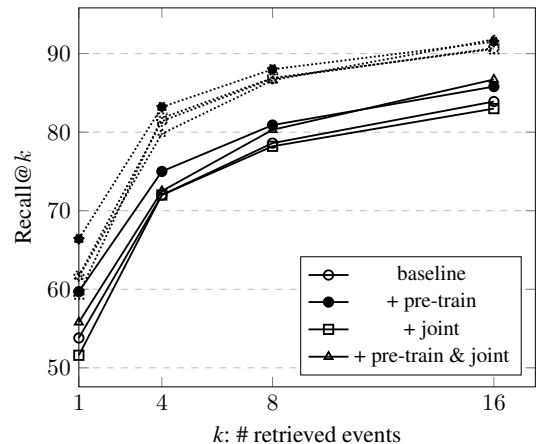

%% file: tables/bi_cr_final.tex
\begin{table*}[t]
\centering
\small
\begin{tabular}{@{}rlccccc@{}}
\toprule
& & Bi-encoder & \multicolumn{4}{c}{Cross-encoder} \\
\midrule
\multicolumn{2}{l}{Methods} & Recall@\emph{min} & Strict Acc & Strict Acc (Top Min) & Macro F1 & Micro F1 \\
\midrule
\multicolumn{7}{l}{\emph{Multilingual}} \\
\midrule
(a) & Baseline & 65.4 / 54.8 & 34.4 / 37.6 & 57.8 / 59.5  & 56.4 / \textbf{62.8} & 53.0 / 58.3  \\
(b) & + HP & \textbf{71.3} / \textbf{58.1} & 40.4 / 39.2 & 61.7 / 60.3 & 60.8 / 62.0  & 57.5 / \textbf{58.7} \\
(c) & + HJL & 63.3 / 51.4 & \textbf{43.6} / \textbf{40.2} & \textbf{63.6} / \textbf{60.8} & \textbf{62.1} / 60.1 &  \textbf{59.2} / 57.5 \\
(d) & + HP + HJL & 67.6 / 55.2 & 38.2 / 39.9 & 60.4 / 60.6 & 57.6 / 61.4 & 54.7 / 57.8 \\
\midrule
\multicolumn{7}{l}{\emph{Crosslingual}} \\
\midrule
(a) & Baseline & 53.8 / 32.8 & 8.5 / 11.9 & 21.2 / 27.5 & 22.1 / 28.8 & 23.6 / 29.2 \\
(b) & + HP & \textbf{59.7} / 37.0 & 8.6 / 10.9 & 18.6 / 25.7 & 22.3 / 28.0  & 24.0 / 28.9 \\
(c) & + HJL & 51.6 / 33.3 & \textbf{9.7} / 12.0 & 22.3 / 26.1   & 21.6 /  28.1 &  23.4 / 29.4 \\
(d) & + HP + HJL & 55.8 / \textbf{38.8} & 9.6 / \textbf{13.1} & \textbf{23.0} / \textbf{28.0} & \textbf{25.7} / \textbf{34.3} & \textbf{27.3} / \textbf{33.1} \\
\bottomrule
\end{tabular}
\caption{Bi-encoder and Cross-encoder performance on multilingual and crosslingual event linking (dev/test). Strict Acc (Top Min) refers to the cross-encoder strict accuracy under Top Min, which is directly comparable to the bi-encoder Recall@\emph{min}.}
\label{tab:bi_cr_final}
\end{table*}

%% file: tables/hd.tex
\begin{table}[t]
\centering
\resizebox{0.45\textwidth}{!}{
\begin{tabular}{@{}lc@{ \ }c@{ \ }c@{ \ }c@{}}
\toprule
& R@1 & R@4 & R@8 & R@16 \\
\midrule 
Multilingual & 46.0 / 45.1 & 69.0 / 72.6 & 76.6 / 81.3 & 79.6 / 84.6 \\
Crosslingual & 52.0 / 37.4 & 78.3 / 60.5 & 86.2 / 75.6 & 90.8 / 83.9 \\
\bottomrule
\end{tabular}
}
\caption{Hierarchical relation extraction results (dev/test) with the top-4 retrieval predictions by the best performing bi-encoder.} 
\label{tab:hd}
\end{table}

%% file: tables/wn_final.tex
\begin{table}[t]
\centering
\small
\begin{tabular}{@{}rlcc@{ \ }c@{ \ }c@{}}
\toprule
& & Bi-encoder & \multicolumn{3}{c}{Cross-encoder} \\
\midrule
\multicolumn{2}{l}{Methods} & R@\emph{min} & Strict Acc & Macro F1 & Micro F1 \\
\midrule
\multicolumn{5}{l}{\emph{Multilingual}}\\
\midrule
(a) & Baseline & \textbf{68.6} & 51.7 & \textbf{67.2} & 62.0  \\
(c) & + HJL & 67.4 & \textbf{55.8} & 65.3 & \textbf{62.7} \\
\midrule
\multicolumn{5}{l}{\emph{Crosslingual}}\\
\midrule
(a) & Baseline & 51.2 & 15.3 & 29.8 & 30.0 \\
(d) & + HP + HJL & \textbf{53.7} & \textbf{21.1} & \textbf{37.6} & \textbf{35.7} \\

\bottomrule
\end{tabular}
\caption{Bi-encoder and cross-encoder performance on multilingual \& crosslingual event linking on Wikinews Dataset}
\label{tab:bi_cr_wn_final}
\end{table}

%% file: plots_final/ml_bi_dev_frac.tex
\begin{figure}[t]
\pgfplotstableread{plots_final/ml_bi_dev_frac.tsv}\loadedtable
\resizebox{0.4\textwidth}{!}{
    \centering
    \begin{tikzpicture}
        \begin{axis}[
            xlabel={\textit{k}: \# retrieved events},
            ylabel={Recall@$k$},
            legend entries={baseline, + pre-train, + joint, + pre-train \& joint},
            legend pos=south east,
            legend style={font=\small},
            xmin=1,
            xtick={1,4,8,16},
            ymajorgrids=true,
            grid style=dashed,
            ]
        
        \addplot[mark=o,mark size=2,thick] table[x=k,y=b]
            {\loadedtable};
        \addplot[mark=*,mark size=2,thick] table[x=k,y=b-hp]
            {\loadedtable};
        \addplot[mark=square,mark size=2,thick] table[x=k,y=b-hjl]
            {\loadedtable};
        \addplot[mark=triangle,mark size=2,thick] table[x=k,y=b-hp-hjl]
            {\loadedtable};

        \end{axis}
    \end{tikzpicture}}
    \caption{Multilingual bi-encoder Recall@$k$ (fraction) on the dev set. }
    \label{fig:ml_bi_dev_frac}
\end{figure}
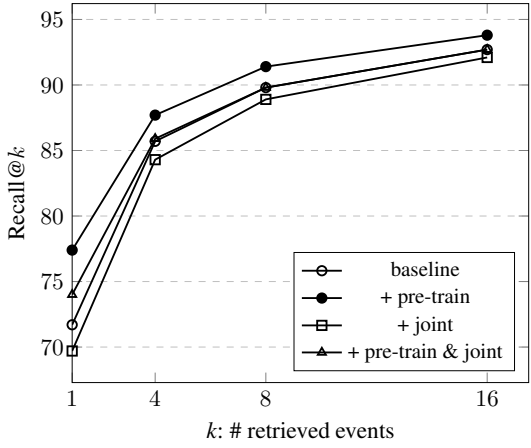

%% file: plots_final/xl_bi_dev_frac.tex
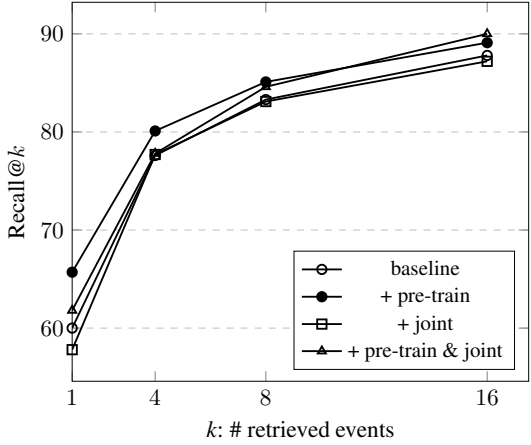
\begin{figure}[t]
\pgfplotstableread{plots_final/xl_bi_dev_frac.tsv}\loadedtable
\resizebox{0.4\textwidth}{!}{
    \centering
    \begin{tikzpicture}
        \begin{axis}[
            xlabel={\textit{k}: \# retrieved events},
            ylabel={Recall@$k$},
            legend entries={baseline, + pre-train, + joint, + pre-train \& joint},
            legend pos=south east,
            legend style={font=\small},
            xmin=1,
            xtick={1,4,8,16},
            ymajorgrids=true,
            grid style=dashed,
            ]
        
        \addplot[mark=o,mark size=2,thick] table[x=k,y=b]
            {\loadedtable};
        \addplot[mark=*,mark size=2,thick] table[x=k,y=b-hp]
            {\loadedtable};
        \addplot[mark=square,mark size=2,thick] table[x=k,y=b-hjl]
            {\loadedtable};
        \addplot[mark=triangle,mark size=2,thick] table[x=k,y=b-hp-hjl]
            {\loadedtable};

        \end{axis}
    \end{tikzpicture}}
    \caption{Crosslingual bi-encoder Recall@$k$ (fraction) on the dev set. }
    \label{fig:xl_bi_dev_frac}
\end{figure}

%% file: tables/ml_per-lang.tex
\begin{table*}[t]
\centering
\small
\begin{tabular}{@{}lccccc@{}}
\toprule
& & Bi-encoder & \multicolumn{3}{c}{Cross-encoder} \\
\midrule
Languages & \# mentions & Recall@$min$ & Strict Acc & Macro F1 & Micro F1 \\
\midrule
\multicolumn{6}{c}{\emph{Multilingual}} \\
\midrule
Afrikaans & 360 & 86.9 & 71.1 & 80.1 & 78.1 \\
Arabic & 1776 & 59.5 & 38.2 & 61.2 & 56.1 \\
Belarusian & 612 & 65.2 & 51.3 & 75.8 & 73.0 \\
Bengali & 136 & 73.5 & 77.2 & 84.4 & 81.9 \\
Bulgarian & 1918 & 48.4 & 45.5 & 75.1 & 69.5 \\
Catalan & 1312 & 52.5 & 40.2 & 61.5 & 59.0 \\
Chinese & 980 & 55.2 & 51.1 & 75.9 & 69.6 \\
Czech & 2051 & 63.0 & 54.8 & 73.6 & 71.4 \\
Danish & 575 & 75.0 & 60.2 & 77.1 & 75.1 \\
Dutch & 2030 & 44.5 & 35.7 & 50.8 & 50.9 \\
English & 10065 & 41.2 & 30.7 & 55.7 & 53.2 \\
Finnish & 2956 & 53.9 & 46.1 & 58.0 & 58.3 \\
French & 6760 & 48.3 & 30.3 & 46.3 & 45.3 \\
German & 8888 & 50.6 & 39.7 & 56.1 & 52.1 \\
Greek & 2841 & 39.8 & 43.0 & 68.3 & 65.4 \\
Hebrew & 1866 & 64.4 & 49.7 & 66.0 & 64.0 \\
Hindi & 91 & 81.3 & 73.6 & 79.3 & 77.9 \\
Hungarian & 974 & 59.4 & 40.9 & 56.9 & 56.3 \\
Indonesian & 678 & 57.4 & 50.4 & 73.8 & 70.4 \\
Italian & 3842 & 44.6 & 29.7 & 47.3 & 46.2 \\
Japanese & 2768 & 54.0 & 46.3 & 62.3 & 57.6 \\
Korean & 897 & 56.1 & 59.5 & 78.2 & 73.0 \\
Malay & 190 & 74.7 & 63.7 & 81.5 & 77.5 \\
Malayalam & 30 & 70.0 & 73.3 & 78.3 & 77.4 \\
Marathi & 16 & 43.8 & 62.5 & 66.6 & 68.3 \\
Norwegian & 895 & 70.2 & 62.5 & 75.5 & 72.9 \\
Persian & 608 & 65.3 & 56.9 & 68.7 & 66.6 \\
Polish & 4749 & 44.8 & 34.3 & 49.3 & 49.5 \\
Portuguese & 1851 & 58.8 & 45.2 & 69.5 & 66.0 \\
Romanian & 1329 & 60.9 & 45.5 & 69.4 & 65.7 \\
Russian & 13221 & 48.8 & 34.8 & 58.5 & 56.1 \\
Serbian & 2770 & 45.2 & 39.7 & 68.5 & 64.5 \\
Slovak & 561 & 60.8 & 57.6 & 72.0 & 68.4 \\
Slovenian & 456 & 66.7 & 57.0 & 67.2 & 67.0 \\
Spanish & 3757 & 56.2 & 40.8 & 54.7 & 53.8 \\
Swahili & 3 & 100.0 & 66.7 & 66.7 & 66.7 \\
Swedish & 1135 & 71.0 & 53.6 & 66.8 & 65.4 \\
Tamil & 36 & 83.3 & 66.7 & 80.9 & 76.5 \\
Thai & 275 & 87.6 & 75.3 & 84.3 & 82.0 \\
Turkish & 1586 & 43.2 & 41.4 & 66.6 & 61.1 \\
Ukrainian & 3551 & 63.8 & 53.1 & 73.5 & 70.6 \\
Vietnamese & 533 & 61.2 & 41.8 & 63.9 & 62.5 \\

\bottomrule
\end{tabular}
\caption{Bi-encoder and cross-encoder performance on multilingual event linking (test set performance) across all languages in the dataset.}
\label{tab:ml_per_lang}
\end{table*}

%% file: tables/xl_per-lang.tex
\begin{table*}[t]
\centering
\small
\begin{tabular}{@{}lccccc@{}}
\toprule
& & Bi-encoder & \multicolumn{3}{c}{Cross-encoder} \\
\midrule
Languages & \# mentions & Recall@$min$ & Strict Acc & Macro F1 & Micro F1 \\
\midrule
\multicolumn{6}{c}{\emph{Crosslingual}} \\
\midrule
Afrikaans & 360 & 67.2 & 34.2 & 40.6 & 39.1 \\
Arabic & 1776 & 24.3 & 6.2 & 20.7 & 19.5 \\
Belarusian & 612 & 36.4 & 9.5 & 31.5 & 30.6 \\
Bengali & 136 & 5.9 & 0.7 & 4.2 & 5.4 \\
Bulgarian & 1918 & 42.3 & 9.1 & 32.6 & 32.2 \\
Catalan & 1312 & 41.8 & 11.8 & 37.5 & 36.7 \\
Chinese & 980 & 23.1 & 7.5 & 22.9 & 21.0 \\
Czech & 2051 & 39.3 & 11.4 & 28.7 & 28.6 \\
Danish & 575 & 48.9 & 23.8 & 44.7 & 40.6 \\
Dutch & 2030 & 32.7 & 14.0 & 33.7 & 33.2 \\
English & 10065 & 41.8 & 18.2 & 49.5 & 45.7 \\
Finnish & 2956 & 33.9 & 16.2 & 33.1 & 32.3 \\
French & 6760 & 34.8 & 13.5 & 33.9 & 32.9 \\
German & 8888 & 38.7 & 13.9 & 33.5 & 31.3 \\
Greek & 2841 & 36.4 & 10.2 & 27.0 & 26.8 \\
Hebrew & 1866 & 32.1 & 8.4 & 20.5 & 20.6 \\
Hindi & 91 & 13.2 & 5.5 & 8.2 & 7.4 \\
Hungarian & 974 & 38.5 & 10.2 & 22.4 & 23.3 \\
Indonesian & 678 & 44.0 & 15.9 & 41.6 & 40.8 \\
Italian & 3842 & 39.0 & 12.9 & 34.5 & 33.5 \\
Japanese & 2768 & 21.2 & 8.3 & 20.9 & 19.8 \\
Korean & 897 & 31.5 & 19.1 & 36.0 & 31.8 \\
Malay & 190 & 44.2 & 12.6 & 34.1 & 33.4 \\
Malayalam & 30 & 16.7 & 6.7 & 12.5 & 13.0 \\
Marathi & 16 & 12.5 & 0.0 & 3.1 & 4.9 \\
Norwegian & 895 & 44.6 & 21.7 & 40.5 & 38.6 \\
Persian & 608 & 35.0 & 7.6 & 22.3 & 21.6 \\
Polish & 4749 & 32.6 & 8.6 & 27.0 & 26.4 \\
Portuguese & 1851 & 44.4 & 17.7 & 44.2 & 40.8 \\
Romanian & 1329 & 45.6 & 14.2 & 36.2 & 34.8 \\
Russian & 13221 & 41.9 & 12.1 & 36.7 & 35.3 \\
Serbian & 2770 & 39.3 & 8.3 & 28.9 & 29.2 \\
Slovak & 561 & 46.0 & 10.0 & 26.8 & 26.3 \\
Slovenian & 456 & 47.1 & 14.2 & 29.9 & 30.2 \\
Spanish & 3757 & 43.3 & 15.3 & 35.9 & 35.0 \\
Swahili & 3 & 0.0 & 0.0 & 11.1 & 15.4 \\
Swedish & 1135 & 48.3 & 18.1 & 35.4 & 34.6 \\
Tamil & 36 & 16.7 & 0.0 & 14.0 & 15.4 \\
Thai & 275 & 44.4 & 20.4 & 42.9 & 40.0 \\
Turkish & 1586 & 36.1 & 10.1 & 25.9 & 26.0 \\
Ukrainian & 3551 & 53.6 & 12.2 & 34.0 & 33.4 \\
Vietnamese & 533 & 41.7 & 12.8 & 37.5 & 33.9 \\

\bottomrule
\end{tabular}
\caption{Bi-encoder and cross-encoder performance on crosslingual event linking (test set performance) across all languages in the dataset.}
\label{tab:xl_per_lang}
\end{table*}